\documentclass[journal=jacsat,manuscript=article]{achemso}

\usepackage[version=3]{mhchem} 
\usepackage[table,xcdraw]{xcolor}
\usepackage{enumerate}
\usepackage{amsfonts}

\author{Tiago Botari}
\affiliation[usp]{Institute of Mathematics and Computer Sciences, University of S\~ao Paulo, S\~ao Carlos - SP, Brazil}
\email{tiagobotari@gmail.com}
\author{Rafael Izbicki}
\affiliation[ufscar]
{Federal University of S\~ao Carlos, S\~ao Carlos - SP, Brazil}
\email{rafaelizbicki@gmail.com}
\author{Andre C. P. L. F. de Carvalho}
\affiliation[usp]{Institute of Mathematics and Computer Sciences, University of S\~ao Paulo, S\~ao Carlos - SP, Brazil}
\email{andre@icmc.usp.br}

\title{Local Interpretation Methods to Machine Learning Using
the Domain of the Feature Space}
\abbreviations{}

\begin{document}

\maketitle          

\begin{abstract}
As machine learning becomes an important part of many real world applications affecting human lives, new requirements, besides high predictive accuracy, become important. One important requirement is transparency, which has been associated with model interpretability. Many machine learning algorithms induce models difficult to interpret, named black box. Black box models are difficult to validate. Moreover,  people have difficulty to trust models that cannot be explained.
Explainable artificial intelligent is an active research area. In particular for machine learning, many groups are investigating new methods able to explain black box models. 
These methods usually look inside the black models to explain their inner work.
By doing so, they allow the interpretation of the decision making process used by black box models.
Among the recently proposed model interpretation methods, there is a group, named local estimators, which are designed to explain how the label of  particular instance is predicted.
For such, they induce interpretable models  on the neighborhood of the instance to be explained. Local estimators  have been successfully used to 
explain specific  predictions. 
Although they provide some degree of model interpretability, it is still not clear what is the best way to implement and apply them.
Open questions include: how to best define the neighborhood of an instance? How to control the trade-off between the accuracy of the interpretation method and its interpretability? How to make the obtained solution robust to small variations on the instance to be explained? To answer to these questions,
we propose and investigate two strategies: (i) using data instance properties to provide improved explanations, and (ii) making sure that the neighborhood of an instance is properly defined by taking the geometry of the domain
of the feature space into account. We evaluate these strategies in a regression task and present experimental results that show that they 
can improve local explanations.

\end{abstract}

\section{Introduction}

Machine learning (ML) algorithms have shown
high predictive capacity for model inference in several application domains. This is mainly due to recent technological advances, increasing number and size of public dataset repositories, and development of powerful frameworks for ML experiments \cite{Dua:2019,imagenet_cvpr09,OpenML2013,tensorflow2015-whitepaper,scikitlearn,paszke2017automatic} 
Application domains where ML algorithms have been successfully used include
image recognition \cite{szegedy2015going}, natural language processing  \cite{xu2015show} and speech recognition \cite{lecun2015deep}. In many of these applications, the safe use of machine learning models and the users' right  to know how decisions affect their life make the interpretability of the models a very important issue. Many currently used machine learning algorithms induce models difficult to interpret and understand how they make decisions, named black boxes.
 
This occurs because several algorithms produce highly complex models in order to better describe the patterns in a dataset.

Most ML algorithms with high predictive performance induce black box models, leading to inexplicable decision making processes. Black box models reduce the confidence of practitioners in the model predictions, which can be a obstacle in many real world applications, such as medical diagnostics \cite{caruana2015intelligible}, science, autonomous driving \cite{bojarski2016end}, and others sensitive domains. In these applications, it is therefore important that predictive models are easy to interpret.

To overcome these problems, many methods that are able to improve model interpretation have been recently proposed; see e.g.
\cite{gilpin2018explaining,molnar2019} for details. These methods aim at providing further information regarding the predictions obtained from predictive models. In these methods, interpretability can occur at different levels: (i) on the dataset; (ii) after the model is induced; and (iii) before the model is induced \cite{lipton2016mythos}. We will focus our discussion on methods for model interpretability that can be applied after the induction of a predictive model by a ML algorithm; these are known as \emph{agnostic methods}.

\begin{figure} [t]
     \centering
     \includegraphics[width=\textwidth]{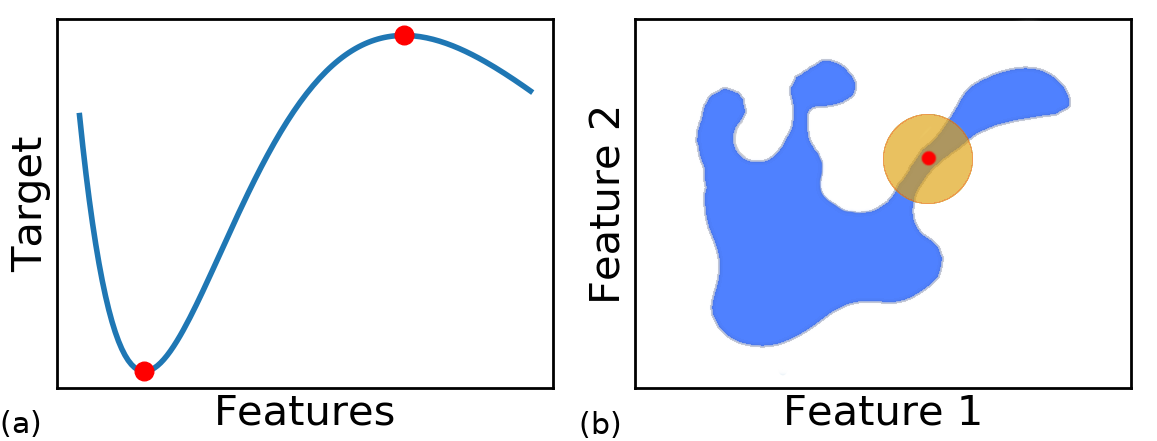}
     \caption{(a) An example where a linear regression of the original features would provide little information regarding the model prediction. The blue continuous line represents the predictive model output as a function of the input, and the red circles represent two critical points of the curve. A local linear regression of the original feature space will produce a limited explanation in the neighborhood of the two critical points. (b) Representation of a domain of a two-dimensional feature problem where the plane defined by the two features is not fully covered. A local sampling can be used to create explanations on the neighborhood of the instance (red circle) that belongs to the correct task domain (blue region) (i.e., the intersection of the orange circle with the blue region) rather than on the orange circle.}
    \label{fig:figure1}
\end{figure}

Model-agnostic interpretation methods are a very promising approach to solve the problem of trust and to uncover the full potential of ML algorithms. These methods can be applied to explain predictions made by models induced by \emph{any} ML algorithm. 
Some well known model-agnostic interpretation methods are described in \cite{friedman2001greedy,fisher2018all,lime,lundberg2016unexpected,vstrumbelj2014explaining}. Perhaps the most well known interpretation method is LIME \cite{lime}, which allows local explanations for classification and regression models.
LIME has been shown to present a very good capability to create local explanations. As a result, LIME has been used to interpret models induced by ML algorithms in different application domains. However, it it still not clear how to make some decisions when implementing and applying LIME and related methods. Some questions that arise are:
\begin{enumerate}[i] 
\item How to best define the neighborhood of an instance?
\item How to control the trade-off between the accuracy of the interpretation model and its interpretability? 
\item How to make the obtained solution robust to small variations on the instance to be explained?
\end{enumerate}

A good local explanation for a given instance $\mathbf{x}^*$ needs to have high fidelity to the model induced by a ML algorithm in the neighborhood of $\mathbf{x}^*$. Although this neighborhood is typically defined in terms of Euclidean 
distances, ideally it should be supported by the dataset.
Thus, the sub-domain used to fit the local explanation model (\textit{i.e.}, a model used to explain the black box model) should reflect the domain where the black model model was induced from. 
For instance, high-dimensional datasets often lie on a submanifold of $\mathbb{R}^d$, in which case defining neighborhoods in terms of the Euclidean distance is not appropriate
\cite{aswani2011regression,lee2016spectral,izbicki2016nonparametric,izbicki2017converting}. To deal with this deficiency, we address issue (i) by creating a technique
that samples training points for the explanation model along the submanifold where the dataset lies on (as opposed to Euclidean neighborhoods). We experimentally show that this technique provides a solution to (iii).

In order to address (ii), we observe that a good local explanation is not necessarily a direct map of the feature space. For some cases, the appropriate local description of the explanation lies on specific properties of the instance. These instance properties can be obtained through a transformation of the feature space. 
Thus, we address issue (ii) by creating local explanations on a transformed space of the feature space.
This spectrum of questions should be elaborated by the specialists of the specific application domain.

In this work, we focus on performing these modifications for regression tasks. However, these modifications can be easily adapted for classification tasks. In Section \ref{sec:properties}, we discuss the use of instance properties, how to deal with the trade-off between explanation complexity and the importance of employing a robust method as an explanatory model. In Section \ref{sec:neigh}, we describe how to improve the local explanation method using the estimation of the domain of feature space. In Section \ref{sec:experiments}, we apply our methodology to a toy example. Finally, Section \ref{sec:conclusions} presents the main conclusions from our work and describes possible future directions.

\section{Model Interpretation Methods}

\subsection{Local Explanation Through Instance Properties}
\label{sec:properties}

A crucial aspect for providing explanations to predictive models induced by ML algorithms is the relevant information to the specific knowledge domain. In some cases, a direct representation of the original set of features of an instance does not reflect the best local behavior of a prediction process. Hence, other instance properties can be used to create clear decision explanations. These properties can be generated through a map of the original features space, \textit{i.e.}, a function of the input  $\mathbf{x}$. Moreover, these instance properties can increase the local fidelity of the explanation with the predictive model.
This can be easily verified when the original feature space is highly limited and providing poor information on the neighborhood of a specific point. This case is illustrated by Figure \ref{fig:figure1} (a).

In order to provide a richer environment to obtain a good explanation, the interpretable model should be flexible to possible questions that an user want to instigate the ML model. Given that the possible explanations are mapped using specific functions of the feature space, we can create an interpretable model using
\begin{equation}
    g(\textbf{x}) = \alpha_0 + \sum_{i=1}^N \alpha_i f_i(\textbf{x})
\end{equation}
where $\textbf{x}$ represents the original vector of features, $\alpha_i$ are the coefficients of the linear regression that will be used as an explanation, and $f_i(.)$ are known functions that map $\textbf{x}$ to the properties (that is, questions) that have a meaningful value for explaining a prediction, or that are necessary to obtain an accurate explanation.

Once $f_i$'s are created, the explainable method should choose which of these functions better represent the predictions made by the original model locally. This can be achieved by introducing an $L1$ regularization in the square error loss function.
More precisely, let $h$ be a black-box model induced by a ML algorithm and consider the task of explaining the prediction made by $h$ at a new instance $\mathbf{x}$.
Let $\mathbf{x}_1',\ldots, \mathbf{x}_{M}'$ be a sample generated on a neighborhood of 
$\mathbf{x}$. The local explanation can be found by minimizing (in $\alpha$)
\begin{equation}
\label{eq:min_local}
    \mathbf{L} = \sum_{k=1}^M (h(\textbf{x}_k') - g(\textbf{x}_k'))^2 + \sum_{i=1}^N \lambda_i|\alpha_i|~,
\end{equation}
where the first term is the standard square error between the induced model and the  explanatory model and the second term is the penalization over the explanatory terms. The value of $\lambda_i$ can be set to control the trade-off among the explanatory terms. 
For instance, if some explanatory terms ($f_i$) are more difficult to interpret, then a larger value can be assigned to $\lambda_i$.

In order to set the objective function (Equation \ref{eq:min_local}), one must be able to sample in a neighborhood of $\mathbf{x}$. To keep consistency over random sampling variations on the neighborhood of $\mathbf{x}$, we decided to use a linear robust method that implements the $L_1$ regularization (see \cite{owen2007robust}). This robust linear regression solves some of the problems of instability of local explanations \cite{alvarez2018robustness}.

Additionally, a relevant question is how to define a meaningful neighborhood around $\mathbf{x}$. In the next section we discuss how this question can be answered in an effective way. 

\begin{figure}[!t]
     \centering
     \includegraphics[width=\textwidth]{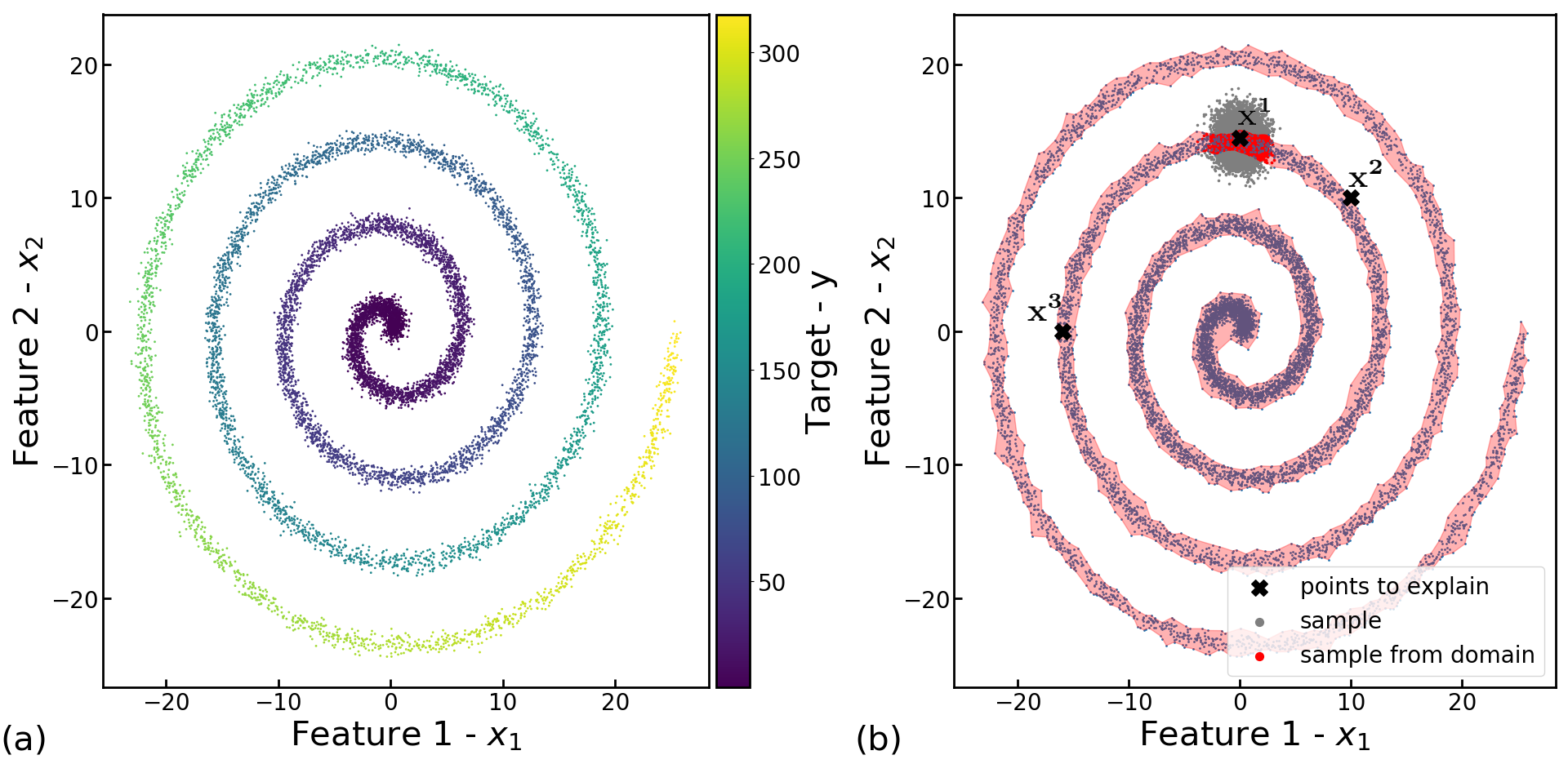}
     \caption{A graphical bi-dimensional representation of the spiral toy model described by Equation \ref{eq:spiral}. (a) Original data where the colors represent the target value ($y$). (b) The domain of feature space (manifold), the blue points represent the original data, the pink polygon is the estimate of the manifold using $\alpha$-shape ($\alpha=1.0$), the black crosses represent the instances to be explained ($\mathbf{x}_{exp}$) (details in Section \ref{points} - $\mathbf{x^1}=(0.0, 14.5)$, $\mathbf{x^2}=(10.0, 10.0)$ and $\mathbf{x^3}=(−16.0, 0.0)$), gray points represent a sample from a normal distribution around the $\mathbf{x}_{exp}$, and the red points correspond to the sample that belong to the estimated domain.}
    \label{fig:spiral_model}
\end{figure}
\subsection{Defining meaningful neighborhoods}
\label{sec:neigh}

\subsubsection{Feature Space}

The training data used by a ML algorithm defines the domain of the feature space. 
In order to obtain a more reliable explanation model, we can use the estimated domain of the feature space for sampling the data needed to obtain this model via Equation \ref{eq:min_local}, $\mathbf{x}_1',\ldots, \mathbf{x}_{M}'$. This approach improves the fidelity and accuracy to the model when compared to standard Euclidean neighborhoods used by other methods \cite{lime}. The estimation of the feature domain is closely related to the manifold estimation problem \cite{wasserman2018topological}. Here, we show how this strategy works by using the $\alpha$-shape technique \cite{edelsbrunner1983shape,edelsbrunner2010alpha} to estimate the domain of the feature space.

\subsubsection{$\alpha$-shape}
The $\alpha$-shape is a formal mathematical definition of the polytope concept of a set of points on the Euclidean space. Given a set of points $S \subset \mathbb{R}^d$ and a real value $\alpha \in [0, \infty)$, it is possible to uniquely define this polytope that enclose $S$. The $\alpha$ value defines an open hypersphere $H$ of radius $\alpha$. For $\alpha \to 0$, $H$ is a point, while for $\alpha \to \infty$, $H$ is an open half-space. Thus, an $\alpha$-shape is defined by all $k$-simplex, $\{k \in \mathbb{Z} | 0 \leq k \leq d\}$, defined by a set of points $s \in S$ where there exist an open hypersphere $H$ that is empty, $H \cap S = \emptyset$, and $\partial H \cap s = s$. In this way, the $\alpha$ value controls the polytope details. For $\alpha \to 0$, the $\alpha$-shape recovered is the set of points $S$ itself, and for $\alpha \to \infty$, the convex hull of the set $S$ is recovered \cite{edelsbrunner1983shape,edelsbrunner2010alpha}.
We define the  neighborhood of an instance $\mathbf{x}$ to be the intersection of an Euclidean ball around $\mathbf{x}$ and the space defined by polytope obtained from the $\alpha$-shape. In practice, we obtain the instances used in Equation \ref{eq:min_local} by sampling new points around $\mathbf{x}$ that belong to the space defined by polytope obtained from the $\alpha$-shape.

\begin{figure}[!t]
     \centering
     \includegraphics[width=\textwidth]{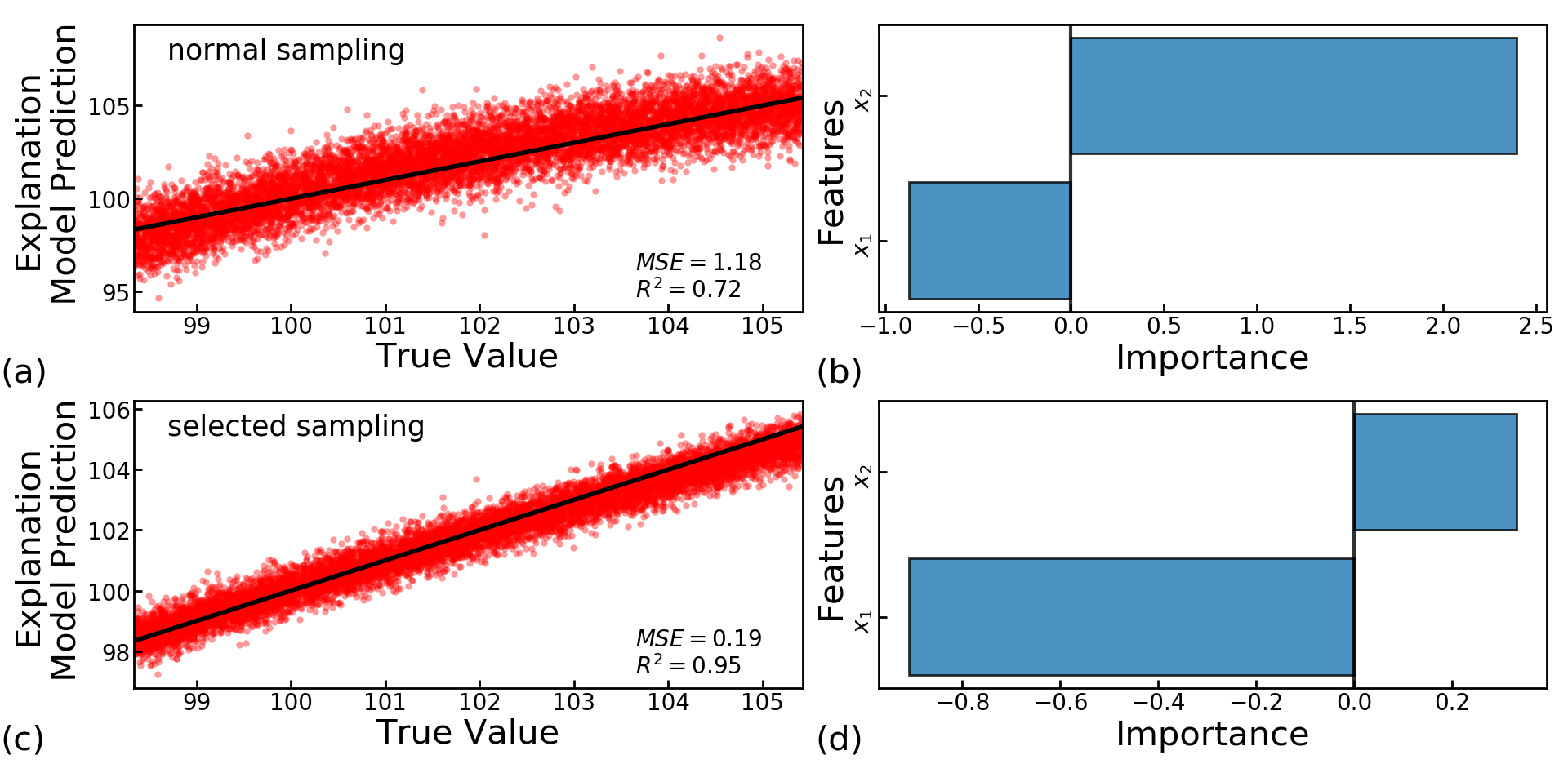}
     \caption{Comparison of prediction performed by the explanation model and the true value of the spiral length using a data set not used during the induction of the model by a ML algorithm. The explanation model was generated for point $\mathbf{x^1}=(0.0, 14.5)$. Figures (a) and (c) show the true label $y$ $versus$ the explanation model prediction. The black line represents the perfect matching between the two values. Figures (b) and (d) show the importance of the features  obtained by the explanation model. Normal sampling strategy ((a) and (b)): $\mbox{MSE}=1.18$; $R^2=0.72$. Selected sampling ((c) and (d)): $\mbox{MSE}=0.19$; $R^2=0.95$.}
    \label{fig:explanation1}
\end{figure}

\begin{figure}[!b]
     \centering
     \includegraphics[width=\textwidth]{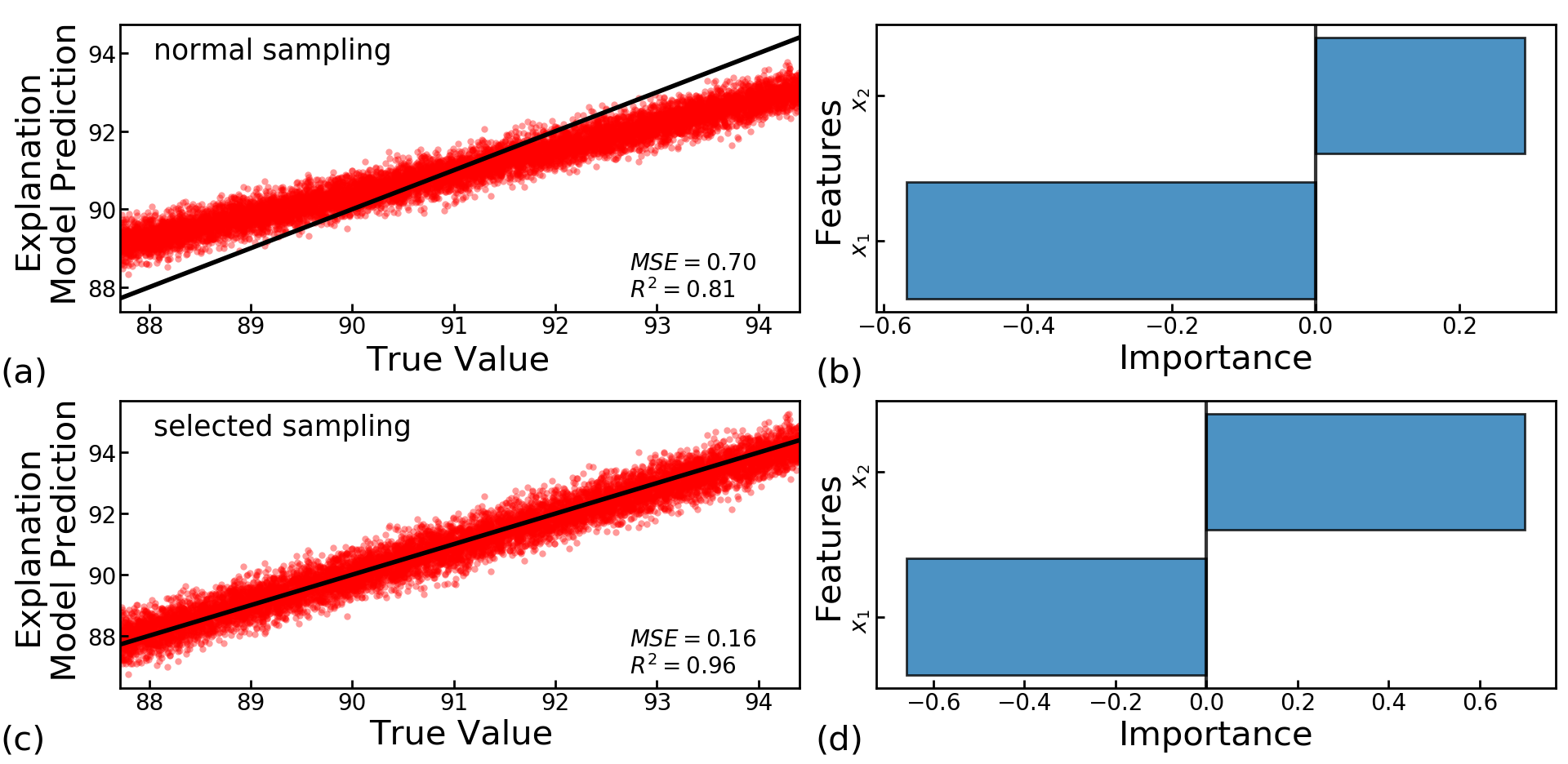}
     \caption{Comparison of prediction performed by the explanation model and the true value of the spiral length using a data set not used during training of the ML model. The explanation model was generated for point $\mathbf{x^2}=(10.0, 10.0)$. Figures (a) and (c) show the true label $y$ $versus$ the explanation model prediction. The black line represents the perfect matching between the two values. Figures (b) and (d) show the importance of the features  obtained by the explanation model. Normal sampling strategy ((a) and (b)): $\mbox{MSE}=0.70$; $R^2=0.81$.
     Selected sampling ((c) and (d)): $\mbox{MSE}=0.16$; $R^2=0.96$.}
    \label{fig:point2}
\end{figure}

\section{Results for a Toy Model: Length of a Spiral}
\label{sec:experiments}
In this section, we present an application of our proposed methodology for a toy model in which the data is generated along a spiral. For such, we use
the Cartesian coordinates of the spiral on the plane as features.

\subsection{Definition}
We explore the toy model described by
\begin{eqnarray}
    x_1 &=& \theta \cos(\theta) + \epsilon_1 ~~~~~~ x_2 = \theta \sin(\theta) + \epsilon_2 \\
    \label{eq:spiral}
    y &=& \frac{1}{2}\left[ \theta \sqrt{1+\theta^2}+ \sinh^{-1} \theta \right] \nonumber
\end{eqnarray}
where $x_1$ and $x_2$ are the values that form the feature vector $\mathbf{x}=(x_1, x_2)$, $\theta$ is a independent variable, $\epsilon_i$, $i \in \{1,2\}$, is a  random noise, and the target value is given by $y$, the  length of the  spiral. This toy model presents some interesting features for our analysis, such as the feature domain over the spiral and the substantial variance of the target value when varying one of the features coordinate while keeping the other one fixed. 

\subsubsection{Instances for Investigation}
\label{points}
We investigate the explanation for 3 specific instances of our toy model: $\mathbf{x^1}=(0.0, 14.5)$, $\mathbf{x^2}=(10.0, 10.0)$ and $\mathbf{x^3}=(-16.0, 0.0)$. For the first point, $\mathbf{x^1}$, we have that the target value (the length of the spiral) will locally depend on the value of $x_1$ , and thus  explanation methods should indicate that the most important feature is $x_1$. For the second value, $\mathbf{x^2}$,  the features    $x_1$ and $x_2$ have the same contribution for explaining such target. Finally, for the third point, $\mathbf{x^3}$, the second feature should be the most important feature to explain the target.

\subsubsection{Data Generation:}

Using the model described in Equation \ref{eq:spiral}, we generated  $80$ thousand data points. These data was generated according to $\theta \sim \mbox{Unif}[0, 8 \pi]$, an uniform distribution. The values of random noise were selected from $\epsilon_1 \sim \mathcal{N}(0, 0.4)$ and $\epsilon_2 \sim \mathcal{N}(0, 0.4)$, where $\mathcal{N}(\mu, \sigma)$  is a normal distribution with mean $\mu$ and standard deviation $\sigma$. The feature space and the target value are shown in Figure \ref{fig:spiral_model} (a). The generated data was split into two sets in which $90\%$ used for training and $10\%$ for testing. Additionally, we test the explanation methods by sampling three sets of data in the neighborhoods of $\mathbf{x^1}$, $\mathbf{x^2}$, and $\mathbf{x^3}$.

\subsubsection{Model induction using a ML algorithm:}

We used a decision tree induction algorithm (DT) in the experiments.
We used the  Classification and Regression Trees  (CART) algorithm implementation 
provided by the scikit-learn \cite{scikitlearn} library.
The model induced by this algorithm using the previously described dataset had as predictive performance $MSE=24.00$ and $R^2=0.997$.

\subsubsection{Determining the $\alpha$-shape of the data:}
For this example, we applied the $\alpha$-shape technique using $\alpha=1.0$. The value of $\alpha$ can be optimized for the specific dataset at hand; see \cite{edelsbrunner2010alpha} for details. The estimation of the domain using the $\alpha$-shape is illustrated by Figure \ref{fig:spiral_model} (b).

\begin{figure}[!t]
     \centering
     \includegraphics[width=\textwidth]{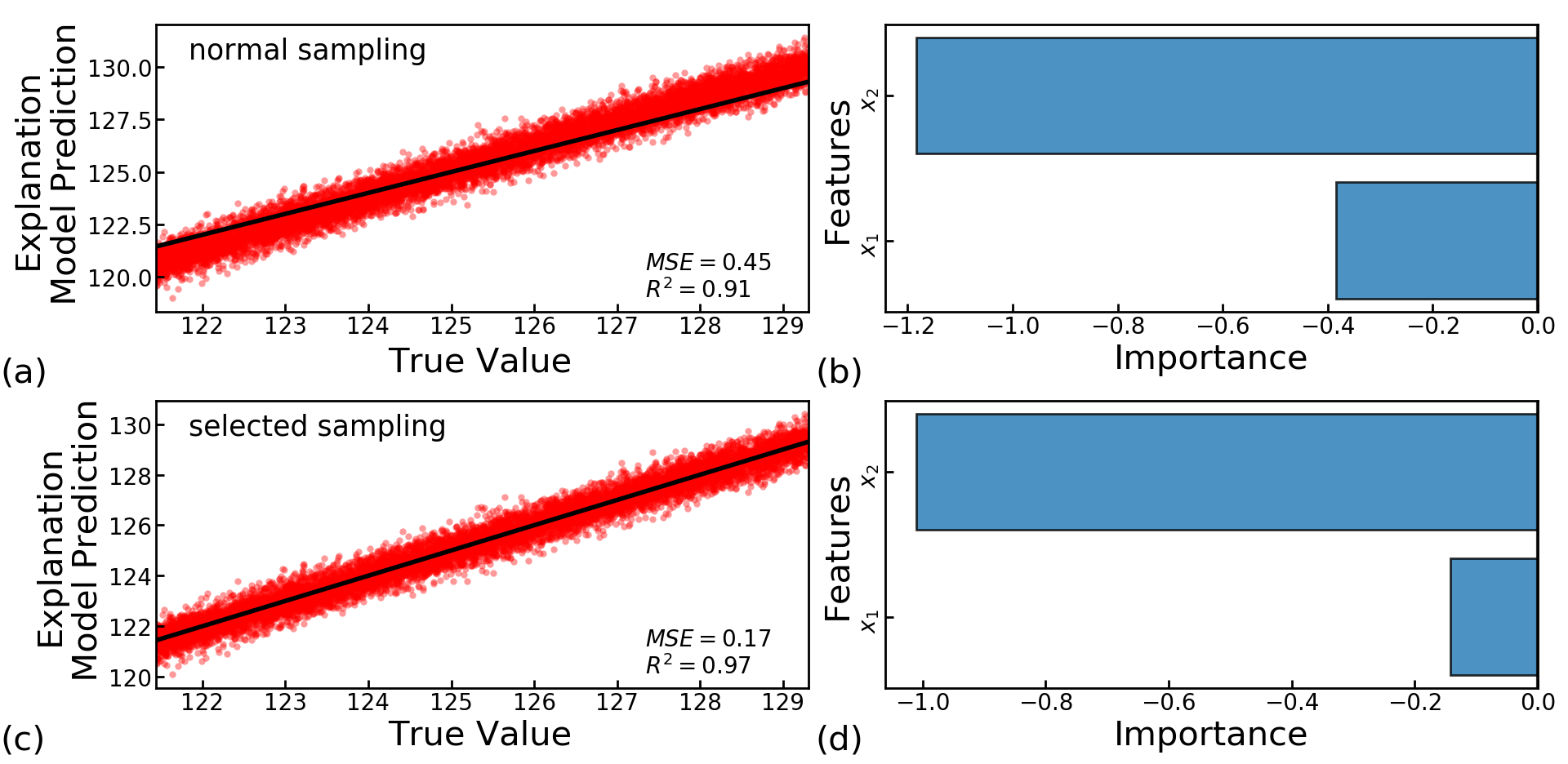}
     \caption{Comparison of prediction obtained by the explanation model and the true value of the spiral length using a data set not used during training of the ML model. The explanation model was generated for point $\mathbf{x^3}=(-16.0, 0.0)$. Figures (a) and (c) show the true label $y$ $versus$ the explanation model prediction. The black line represents the perfect matching between the two values.  Figures (b) and (d) show the importance of the features  obtained by the explanation model.
     Normal sampling strategy ((a) and (b)): $\mbox{MSE}=0.45$; $R^2=0.91$.
     Selected sampling ((c) and (d)):
     $\mbox{MSE}=0.17$; $R^2=0.97$.}
    \label{fig:point3}
\end{figure}
\subsection{Local Explanation}
The local explanation was generated though a linear regression fitted to a data generated over the neighborhood of the point for which the explanation was requested ($\mathbf{x_{exp}}$). We use the linear robust method  available on the scikit-learn package \cite{scikitlearn}.  

\subsubsection{Explanation for instance $\mathbf{x^1}=(0.0, 14.5)$:} The obtained explanation using the standard sampling approach
(hereafter \emph{normal sampling}) presents low agreement with true value of the spiral length (Figure \ref{fig:explanation1}(a)). We also noticed that this explanation is unstable with respect to sampling variations (even though we use a robust method to create the interpretation), and indicates that the best feature to explain the ML algorithm locally is  $x_2$ (Figure \ref{fig:explanation1}(b)). This description is inaccurate (see discussion in Section \textbf{Instances for Investigation}). On the other hand, when the sampling strategy is performed over the correct domain of the feature space (hereafter \emph{selected sampling}), we obtain an explanation method with high predictive accuracy (i.e., that accurately reproduces the true target value - Figure \ref{fig:explanation1}(c)). Moreover, the feature that best explains such prediction is $x_1$ (Figure \ref{fig:explanation1}(d)), which is in agreement with our expectation. 

\subsubsection{Explanation for instances $\mathbf{x^2}=(10.0, 10.0)$ and $\mathbf{x^3}=(−16.0, 0.0)$:} 
We also analyzed the other two points to demonstrate the capability of the selected sampling to capture the correct feature importance. For the instance $\mathbf{x^2}$, the features importance is almost equally divided between the two features (Figure \ref{fig:point2}). For the instance $\mathbf{x^3}$, the most important feature is $x_2$, with importance of $-1.0$ (figure \ref{fig:point3}). In the case of $\mathbf{x^3}$, the normal sampling strategy produced a good explanation (figure \ref{fig:point3}(b)). However, we noticed that this result is unstable due to random variation in the sampling. All results presented here are in agreement with our discussion in Section \textbf{Instances for Investigation}. 

\subsection{Robustness of Explanations}

Good explanation models  for  $\mathbf{x*}$ should be stable to small perturbations around $\mathbf{x*}$. To illustrate the stability of our method, we generated explanations for instances in the neighborhood of $\mathbf{x^1}$: $\mathbf{x^{1a}}=(-2.0, 14.5)$, $\mathbf{x^{1b}}=(1.0, 14.0)$ and $\mathbf{x^{1c}}=(0.5, 13.7)$.
Table \ref{tab:robustness} shows that the explanations created for these points using selected sampling are compatible with those for $\mathbf{x^1}$. On the other hand, the normal sampling strategy is unstable. These results demonstrate that using the domain defined by the feature space can improve the robustness of a local explanation of an instance.

\begin{table}[]
\centering
\caption{Local explanations generated for instances around instance $\mathbf{x^1}$: for normal and selected sampling strategies. MSE and R$^2$ measured between true values and predictions performed by the local explanation model.}
\resizebox{1.0\textwidth}{!}{%
\begin{tabular}{p{1.5cm}p{1.5cm}p{1.5cm}p{2.5cm}p{2.5cm}p{1.5cm}l}
\rowcolor[HTML]{FFFFFF} 
\multicolumn{7}{l}{\cellcolor[HTML]{FFFFFF}{\color[HTML]{000000} {\small{Normal Sampling}}}}     \\ \hline
\rowcolor[HTML]{EFEFEF} 
point             & $x_1$ & $x_2$ & $x_1$ Importance & $x_2$ Importance & MSE  & R$^2$ \\ \hline
{$\mathbf{x^1}$}    & 0.0   & 14.5  & -0.92            & 2.46             & 1.18 & 0.72  \\
\rowcolor[HTML]{EFEFEF} 
\textbf{$\mathbf{x^{1a}}$} & -2.0  & 14.5  & -1.07      & 1.87       & 6.19 & 0.64  \\
\textbf{$\mathbf{x^{1b}}$} & 1.0   & 14.0  & -0.89      & 3.91        & 8.99 & 0.46  \\
\rowcolor[HTML]{EFEFEF} 
\textbf{$\mathbf{x^{1c}}$} & 0.5   & 13.7  & -0.95       & 1.47       & 1.09 & 0.93  \\ \hline\hline

\multicolumn{7}{l}{{\small Selected Sampling}}                                                      \\ \hline
\rowcolor[HTML]{EFEFEF} 
point             & $x_1$ & $x_2$ & $x_1$ Importance & $x_2$ Importance & MSE  & R$^2$ \\ \hline
\rowcolor[HTML]{FFFFFF} 
$\mathbf{x^1}$             & 0.0   & 14.5  & -0.96            & 0.33             & 0.19 & 0.95  \\
\rowcolor[HTML]{EFEFEF} 
$\mathbf{x^{1a}}$          & -2.0  & 14.5  & -0.98            & 0.31             & 0.30  & 0.98  \\
\rowcolor[HTML]{FFFFFF} 
$\mathbf{x^{1b}}$          & 1.0   & 14.0  & -0.97            & 0.07             & 0.21 & 0.99  \\
\rowcolor[HTML]{EFEFEF} 
$\mathbf{x^{1c}}$          & 0.5   & 13.7  & -0.96            & 0.39             & 0.39 & 0.99  \\ \hline
\end{tabular}
}
\label{tab:robustness}
\end{table}
\normalsize
\section{Conclusion}
\label{sec:conclusions}

In order to increase trust and confidence on black box models induced by ML algorithms, 
explanation methods must be reliable, reproducible and flexible with respect to the nature of the questions asked. Local agnostic-model explanations  methods have many advantages that are  aligned with these points. Besides, they can be applied to any ML algorithm. However, the standard of the existing agnostic methods present problems in producing reproducible explanation, while maintaining accuracy to the original model.
To overcome these limitations, we developed new strategies to overcome them. For such, the proposed strategies address the following issues: (i) estimation of the domain of the feature space in order to provide meaningful neighborhoods; (ii) use of different penalization level on explanatory terms; and (iii) employment of robust techniques for fitting the explanatory method.  

The estimation of the domain of the features space should be performed and used during the sampling step of local interpretation methods. This strategy increases the accuracy of the local explanation. Additionally, using robust regression methods to create the explainable models is beneficial to obtain stable solutions. However, our experiments show that robust methods are not enough; the data must be sampled   taking the domain of the feature space into account, otherwise the generated explanations can be meaningless.

Future work includes testing other methods for estimating manifolds such as diffusion maps \cite{coifman2006diffusion} and isomaps \cite{tenenbaum2000global}, extending these ideas to classification problems, and  investigating the performance of our approach on real datasets.

\section*{Acknowledgments}

The authors would like to thank CAPES and CNPq (Brazilian Agencies) for their financial support. T.B. acknowledges support by Grant 2017/06161-7, S\~ao Paulo Research Foundation (FAPESP). R. I. acknowledges support by Grant 2017/03363  (FAPESP) and Grant 306943/2017-4 (CNPq). The authors acknowledge Grant 2013/07375-0 - CeMEAI - Center for Mathematical Sciences Applied to Industry from S\~ao Paulo Research Foundation (FAPESP).
T.B. thanks Rafael Amatte Bizao for review and comments.
\bibliographystyle{unsrt}
\bibliographystyle{splncs04}
\bibliography{bib}

\end{document}